%% file: corl-2019.tex
\documentclass{article}

\usepackage[final]{corl_2019} 

\usepackage{graphicx}        
\usepackage{wrapfig}
\usepackage[tight]{subfigure}

\usepackage{caption}
\usepackage[bottom]{footmisc}

\usepackage{amsmath}
\usepackage{amsfonts}
\usepackage{amssymb}
\usepackage{bbold}  
\usepackage{multirow}
\usepackage{mdframed}
\usepackage{bm} 
\usepackage{gensymb} 

\usepackage{tabularx}
\usepackage{booktabs}
\usepackage{multirow}
\usepackage{xspace} 
\usepackage{algorithm}
\usepackage{algorithmic}
\usepackage{wrapfig}

\usepackage{tikz}
\usetikzlibrary{positioning}

\usepackage[normalem]{ulem}

\newcommand{\argmax}[1]{\ensuremath{\underset{#1}{\textrm{arg max}}\;}}

\usepackage[textsize=footnotesize]{todonotes}
\title{Language-guided Semantic Mapping and Mobile Manipulation in Partially Observable Environments}

\author{
  Siddharth Patki\\
  University of Rochester\\
  \texttt{spatki@ur.rochester.edu} \\
  \And
  Ethan Fahnestock\\
  University of Rochester\\
  \texttt{efahnest@u.rochester.edu} \\
  \AND
  Thomas M.\ Howard\\
  University of Rochester\\
  \texttt{thomas.howard@rochester.edu} \\
  \And
  Matthew R.\ Walter\\
  Toyota Technological Institute at Chicago\\
  \texttt{mwalter@ttic.edu} \\
}

\makeatletter
\hypersetup{
  pdftitle={\@title},
  pdfauthor={Siddharth Patki, Ethan Fahnestock, Thomas M. Howard, and Matthew R. Walter}
}
\makeatother

\begin{document}
\maketitle

\input{abstract}


\input{intro}
\input{related}
\input{approach}
\input{experimental-design}
\input{results}
\input{conclusion}

\acknowledgments{This work was supported in part by the National Science Foundation under grants IIS-1638072 and IIS-1637813 and by ARO grant W911NF-15-1-0402.}


\clearpage
\bibliography{references}  
\end{document}

%% file: abstract.tex
\begin{abstract}
    Recent advances in data-driven models for grounded language understanding have enabled robots to interpret increasingly complex instructions. Two fundamental limitations of these methods are that most require a full model of the environment to be known a priori, and they attempt to reason over a world representation that is flat and unnecessarily detailed, which limits scalability. Recent semantic mapping methods address partial observability by exploiting language as a sensor to infer a distribution over topological, metric and semantic properties of the environment. However, maintaining a distribution over highly detailed maps that can support grounding of diverse instructions is computationally expensive and hinders real-time human-robot collaboration. We propose a novel framework that learns to adapt perception according to the task in order to maintain compact distributions over semantic maps. Experiments with a mobile manipulator demonstrate more efficient instruction following in a priori unknown environments.
\end{abstract}

%% file: intro.tex

\section{Introduction} \label{sec:intro}

\begin{wrapfigure}{r}{0.4\textwidth}
  \centering
  \vspace{-4mm}
  \includegraphics[width=0.38\textwidth]{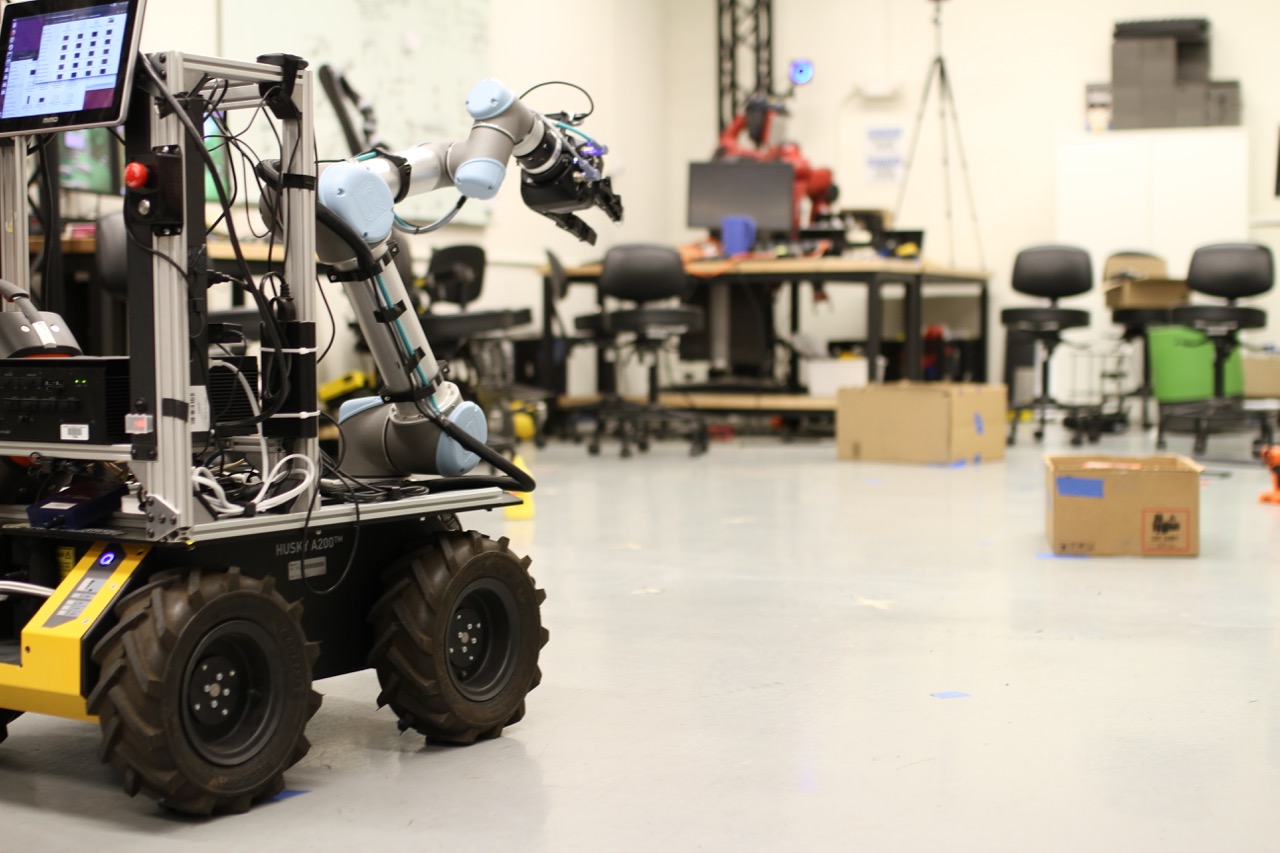}
  \caption{A user commands a robot to ``retrieve the ball inside the box'' in an a priori unknown environment.}
  \label{fig:intro}
  \vspace{-2mm}
\end{wrapfigure}
Realizing robots that can work effectively alongside people in cluttered, unstructured environments (Fig.~\ref{fig:intro}) requires command and control mechanisms that are both intuitive and efficient. Natural language provides a flexible medium through which users can communicate with robots without the need for specialized interfaces or significant training. For example, a voice-controllable wheelchair~\citep{hemachandra11} permits people with limited mobility to independently navigate their environments without using sip-and-puff arrays or head-actuated switches.

\begin{figure}[!tb]
  \centering
  \subfigure[exhaustive world map]{\includegraphics[width=0.32\linewidth]{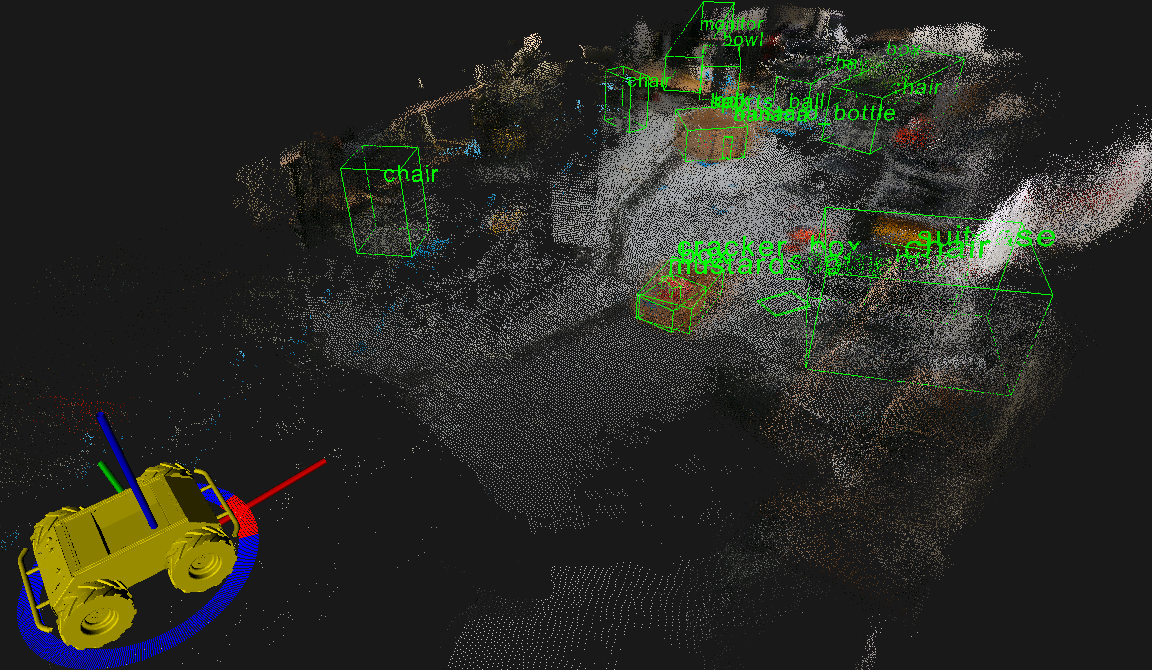}}\hfil%
  \subfigure[distribution of exhaustive maps]{\includegraphics[width=0.32\linewidth]{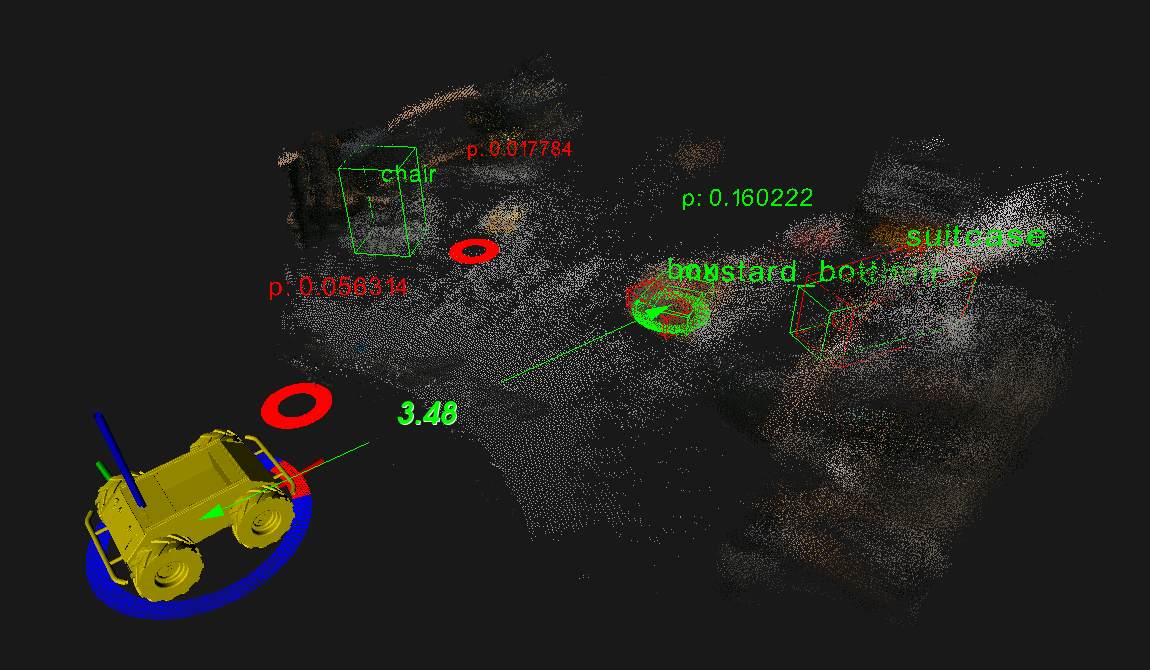}}\hfil%
  \subfigure[distribution of compact maps]{\includegraphics[width=0.32\linewidth]{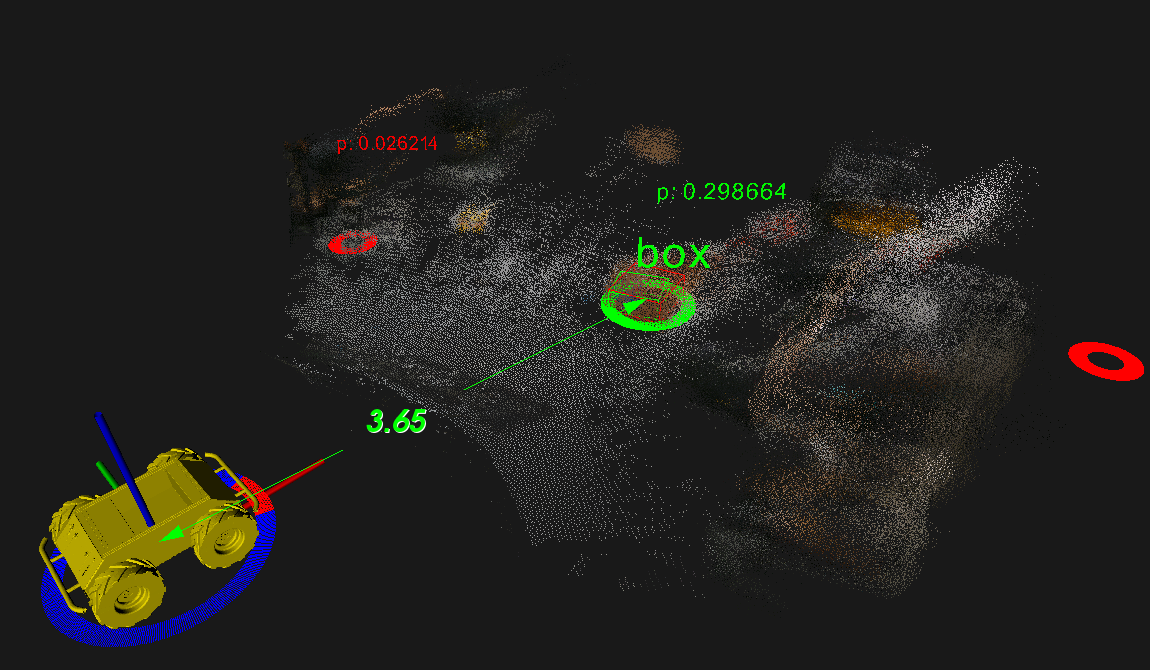}}
  \caption{Our framework learns to exploit environment and task-related information implicit in a given utterance to infer a distribution over compact task-relevant maps in a priori unknown environments. Consider the command ``retrieve the ball inside the box''. Traditional approaches to language grounding involve reasoning over (a) a highly detailed model of the environment that is computationally expensive to maintain and assumed to be known a priori. To enable grounding in unknown or partially observed environments, recent methods consider maintaining a distribution over (b) highly detailed maps that include all observed objects as well as the hypothesized location of unknown objects referenced in the utterance. In contrast, our proposed approach learns to reason over (c) a distribution of compact maps that model only task-relevant objects by adapting perception based on the utterance. In the above figures, circles denote the hypothesized locations for a box that contains a ball from different maps in the distribution.}
  \label{fig:motivation}
\end{figure}

Significant progress in data-driven approaches to language understanding have enabled robots to both interpret and generate complex free-form utterances in a variety of domains~\citep{thomason16, shridhar18, kollar10, matuszek10, matuszek12a, thomason15}. Symbol grounding-based methods formulate language understanding as a problem of associating linguistic phrases with their corresponding referents in the robot's model of its state and action space. This places two fundamental limitations on grounding-based approaches to language understanding. First, most contemporary solutions require a priori knowledge of the robot's environment in the form of a ``world model'' that expresses the metric and semantic properties of every object and location in the robot's environment. This model is typically created by augmenting a SLAM-generated metric map with manually and/or automatically inferred semantic information. Critically, this prevents the robot from interpreting commands in unobserved or partially observed environments. Second, advances in sensor technology and computer vision algorithms have give rise to a wealth of information that can be infused into these world models. This results in exhaustively detailed representations of the environment (Fig.~\ref{fig:motivation}a), such as those that could model all of the spoons on a table, or door handles on the doors and their affordances. While these models are sufficient for language grounding, the computational cost of perceiving all objects and performing symbol grounding in their context~\citep{paul2018efficient} precludes real-time language grounding. Conversely, a poorly detailed model of the environment that assumes a coarse, static representation of objects limits the diversity of the instructions that can be grounded.

Towards addressing the problem of interpreting instructions in a priori unknown environments, recent work by \citet{hemachandra15} presents an approach that exploits information communicated by the human that may indirectly inform a robot about its target environment. Specifically, their approach extracts spatial-semantic properties of objects and regions conveyed as part of an instruction in order to infer a distribution over possible maps, permitting language understanding in novel environments. However, their method maintains a distribution over unnecessarily detailed semantic maps that are generated with the help of fiducials~\citep{olson2011apriltag}. In practice, building such highly detailed world models without using fiducials is computationally expensive, and hinders smooth human-robot collaboration. A recent line of work~\citep{patki18a,patki2019a} proposes a model that learns to dynamically adapt the configuration of the robot's perception pipeline by inferring the classifiers needed to express the symbols that would later be needed by the symbol grounding model. In this work, we present an efficient approach to collaborative mobile manipulation that jointly learns the models for map inference and adaptive perception. The proposed framework infers the subset of perceptual classifiers needed to efficiently update environment models in an online fashion for the execution of multiple tasks in novel or partially observed worlds. Experimental results on a Clearpath Robotics Husky A200 Unmanned Ground Vehicle outfitted with a Universal Robotics UR5 manipulator and Robotiq gripper demonstrate faster task execution in partially observed worlds compared to a fixed-perception baseline for mobile manipulation tasks.

%% file: related.tex
\section{Related Work} \label{sec:related}
Statistical approaches to language understanding have enabled robots to follow complex free-form instructions involving object manipulation~\cite{paul2018efficient, thomason16, shridhar18}, navigation~\cite{kollar10, matuszek10, matuszek12a, thomason15} and mobile manipulation~\cite{tellex11, walter14b}. A common approach to language understanding is to treat it as a symbol grounding problem~\cite{tellex11, paul2018efficient}, whereby one learns a model that associates (i.e., ``grounds'') each word in an utterance to its corresponding referent in the robot's model of its state and action space. Such approaches typically require a ``world model'' to be known a priori in the form of a map that expresses the location, geometry, semantic type, and colloquial name of all objects and regions in the environment. In practice, these maps are often generated by first using a state-of-the-art SLAM algorithm~\cite{walter07, olson06, kaess08}, which produces flat representations that only model spatial information. Semantic and topological properties are then manually added to realize a representation sufficient for language understanding. Notable exceptions include the work of \citet{duvallet13}, which learns to follow navigational instructions in unknown environments based upon human demonstrations, as well as recent work on language-based visual navigation in novel environments~\cite{mei16, anderson18}. The latter differ from our work in that they map language directly to actions, and do not (explicitly) infer a compact world model from language. Meanwhile, statistical parsing-based methods~\cite{matuszek10, chen11, matuszek12a, thomason15} associate natural language utterances to a meaning representation that typically takes the form of a lambda calculus. Such an approach avoids the need for an explicit world model, typically at the expense of requiring a down-stream controller capable of executing inferred plans in unknown environments.

Also relevant is recent work that focuses on grounding unknown or ambiguous utterances. One approach to dealing with ambiguous utterances is to utilize inverse grounding~\cite{tellex14, gong18} to generate targeted questions for the user that are deemed to be most informative, e.g., in terms of the reduction in entropy for the grounding distribution~\cite{tellex12}. Meanwhile, several methods learn a priori unknown grounding models by exploring the relationship between novel linguistic predicates and the robot's world model and/or its percepts~\cite{thomason16, she17, tucker17, thomason18}. Our work differs in that we assume that the concepts are known, but that the instantiations of these concepts in the robot's environment are unknown.

Similar to how our framework performs map inference, state-of-the-art semantic mapping frameworks build rich representations of the world from the robot's multimodal sensor streams~\cite{zender08, pronobis10}, including linguistic descriptions~\cite{walter13, hemachandra14}. The latter methods attempt to reason over all perceptual cues irrespective of the utterance. In contrast, our framework uses natural language as another sensor to maintain a distribution over the metric, topological, and semantic properties of the unknown environment. This distribution is then used for language grounding and planning.

Most language grounding methods perform inference over the entire power set of objects, regions, actions, and other constituents in the search space. The Distributed Correspondence Graph (DCG)~\cite{paul2018efficient} reduces the complexity of grounding from exponential to linear by performing inference separately across conditionally independent constituents in a graphical model of language grounding. Recent variations of the DCG~\citep{paul2018efficient} further improve computational efficiency by performing inference in a multi-stage, coarse-to-fine manner. We leverage DCG in this work to learn the proposed models for adapting perception, map inference, and symbol grounding.

%% file: approach.tex
\section{Technical Approach} \label{sec:approach}
Many contemporary approaches frame natural language understanding as inference over a learned distribution that associates linguistic elements to their corresponding referents in a symbolic representation of the robot's state and action space. The space of symbols $\bm{\Gamma} = \left\{\gamma_1,\gamma_2 ... \gamma_n\right\}$ includes concepts derived from the robot's environment model, such as objects and locations, and includes the viable robot behaviors, such as navigating to a desired location or manipulating a specific object. The distribution over symbols is conditioned on the parse of the utterance $\bm{\Lambda} = \left\{\lambda_1,\lambda_2 ... \lambda_n\right\}$, and a  model of the world $\Upsilon$ that expresses environment knowledge extracted from sensor measurements $\bm{z}_{1:t}$ using a set of perceptual classifiers $\bm{P} = \left\{p_1,p_2 ... p_n\right\}$. Natural language understanding framed as a symbol grounding problem then follows as maximum a posteriori inference over the power set of referent symbols $\mathcal{P}(\bm{\Gamma})$.
\begin{equation}
  \bm{\Gamma^*} = \argmax{\mathcal{P}(\bm{\Gamma})} \; p( \bm{\Gamma}  \vert \bm{\Lambda}, \Upsilon)
  \label{eqn:basic-1}
\end{equation}
This approach reasons in the context of a known model of the world $\Upsilon$ that is assumed to express all information necessary to ground the given utterance. This precludes language understanding in unobserved (i.e., novel) or partially observed environments for which the world model is incomplete, thereby making accurate inference \eqref{eqn:basic-1} infeasible. To address this problem, we instead treat symbol grounding as inference conditioned on a latent model of the robot's environment $\overline{\bm{\Upsilon}}$. Specifically, we learn a model that exploits environmental information implicit in an utterance to build a distribution over the topological, metric, and semantic properties of the environment
\begin{equation}
 p(\overline{\bm{\Upsilon}}_t \vert \bm{\Lambda}_{1:t}, \bm{z}_{1:t}, \bm{u}_{1:t}, \bm{P}),
 \label{eqn:world-distribution}
\end{equation}
where $\bm{\Lambda}_{1:t}$, $\bm{z}_{1:t}$, and $\bm{u}_{1:t}$ denote the history of utterances, sensor observations and odometry, respectively, and $\bm{P}$ is the set of classifiers in the robot's perception pipeline. This allows maintaining a world model distribution that not only embeds the perceived entities from sensor data but also models the unperceived information about the environment expressed in the utterance. This enables symbol grounding in unknown or partially observed environments. As we describe shortly, we maintain this distribution using a Rao-blackwellized particle filter, whereby each particle effectively denotes a hypothesized world model $\Upsilon^{i}_t \in \overline{\bm{\Upsilon}}_t$.

Treating the environment model as a latent random variable, we formulate symbol grounding as a problem of inferring a distribution over robot behaviors $\overline{\bm{\beta}}_t$. A behavior $\beta_t$ is a representation of the intended robot actions expressed by the symbols in the inferred groundings $\bm{\Gamma^*}_t$. Each behavior $\beta^i_t \in \overline{\bm{\beta}}_t$ is inferred in the context of the corresponding world $\Upsilon^{i}_t \in \overline{\bm{\Upsilon}}_t$ and the instruction $\bm{\Lambda}_t$.
The optimal trajectory $\bm{x}_{t}^*$ that the robot should take in the context of a distribution of behaviors then amounts to a planning under uncertainty problem formulated as inference
\begin{equation}
  \bm{x}^*_t = \argmax{\bm{x}_t \in \bm{X}_t} \; \sum_{\Upsilon^{i}_t \in \overline{\bm{\Upsilon}}_t} \underbrace{p(\bm{x}_t \vert \beta_t^{i}, \Upsilon_t^i)}_{\text{path planning}} \times
  \underbrace{p( \beta_t^{i} \vert \bm{\Lambda}_t, \Upsilon_t^i)}_{\text{behavior inference}} \times \underbrace{p( \Upsilon_t^i \vert  \bm{\Lambda}_{1:t}, \bm{z}_{1:t}, \bm{u}_{1:t}, \bm{P})}_{\text{semantic mapping}}
  \label{eqn:basic-2}
\end{equation}
\begin{figure}[!t]
	\centering
	\begin{tikzpicture}[
		ap/.style={rectangle, draw=black, text=white, fill=black!40!orange, line width=0.03cm, minimum size=10mm, text width=2.5cm, align=center},
		annotation/.style={rectangle, draw=black, text=white, fill=black!40!green, line width=0.03cm, minimum size=10mm, text width=2.5cm, align=center},
		behavior/.style={rectangle, draw=black, text=white, fill=black!40!cyan, line width=0.03cm, minimum size=10mm, text width=2.5cm, align=center},
		basic/.style={rectangle, draw=black, fill=white!60, line width=0.03cm, minimum size=6mm, text width=1.5cm},
		on grid, auto,
		scale=0.7, every node/.style={transform shape},
		align=center
		]

		\node[basic]    (human)                             {human};
		\node[basic]    (parser)  [above=0.9in of human]                          {parser};
		\node[inner sep=0,minimum size=0,right=1.0in of parser] (inviz) {};
		\node[annotation]   (nluAnnotation)  [right=1.8 in of parser]               {\textbf{NLU} \\ \textbf{annotation inference}};
		\node[ap]       (nluPercp)  [above=0.9in of nluAnnotation]          {\textbf{NLU} \\ \textbf{perception inference}};
		\node[behavior]    (nluBehavior)     [below=0.9in of nluAnnotation]             {\textbf{NLU} \\ \textbf{behavior inference}};
		\node[ap]           (AP)          [right=2in of nluPercp]                  {adaptive perception};
		\node[annotation]   (semMap)      [right=2in of nluAnnotation]                  {semantic mapping};
		\node[inner sep=0,minimum size=0,below=0.4in of semMap] (inviz2) {};
		\node[behavior]    (policy)      [right=2in of nluBehavior]                  {policy planner};
		\node[basic]        (exec)        [right=1.4in of policy]                  {executive};
		\node[basic]        (robot)       [above=of exec, right=1.4in of semMap]                 {robot};

		\draw [->,line width=1pt] (human.north) -- node [left] {\footnotesize instruction} (parser.south);
		\draw [->,line width=1pt] (inviz.north) |- (nluPercp.west);
		\draw [->,line width=1pt] (inviz.south) |- (nluBehavior.west);
		\draw [->,line width=1pt] (parser.east) -- node [ above, align=center, text width=0.7in, pos=0.4] {\footnotesize parse tree\\ $\bm{\Lambda_t}$} (nluAnnotation.west);

		\draw [->,line width=1pt] (AP.south) -- node [right=-0.2cm, align=center, text width=1in] { \footnotesize compact \\ local world $\Upsilon^*_t$} (semMap.north);
		\draw [->,line width=1pt] (nluPercp.east) -- node [above] {\footnotesize detector \\ \footnotesize symbols $\bm{P}^*_t$} (AP.west);
		\draw [->,line width=1pt] (nluAnnotation.east) -- node [above] {\footnotesize annotation \\ \footnotesize symbols $\bm{\alpha}_t$} (semMap.west);
		\draw [->,line width=1pt] (nluBehavior.east) -- node [above] {\footnotesize distribution over \\ \footnotesize behaviors $\overline{\bm{\beta}}_t$} (policy.west);
		\draw [->,line width=1pt] (semMap.south) -- node [right=-0.3cm,align=center, text width=1in] {\footnotesize distribution over maps $\overline{\bm{\Upsilon}}^{*}_t$} (policy.north);
		\draw [->,line width=1pt] (inviz2.west) -| (nluBehavior.north);

		\draw [->,line width=1pt] (policy.east) -- node [above] {\footnotesize goal $g^*_t$} (exec.west);
		\draw [->,line width=1pt] (exec.north) -- node [right] {\footnotesize commands \\ $\bm{a}_t$} (robot.south);
		\draw [->,line width=1pt] (robot.north) |- node [right, align=center, text width=0.7in, pos=0.25] {\footnotesize observations $\bm{z}_{1:t}, \bm{u}_{1:t}$} (AP.east);

		\end{tikzpicture}
	\caption{The system architecture for language understanding in unknown environments using adaptive perception, semantic mapping, and natural language symbol grounding.  The three language understanding models learned from the corpus of annotated instructions are highlighted in bold.}
	\label{fig:system-architecture}
\end{figure}
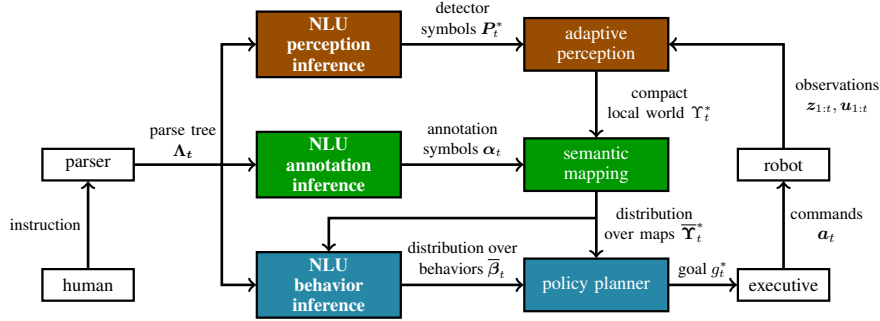
As the robot explores its environment, the distribution over world models is updated by incorporating new detections from the robot's perception pipeline and by incorporating information contained in any instructions that follow. Every time an update is made to the world distribution, the optimal trajectory is recomputed.

The ability to ground diverse natural language instructions is inherently linked to the richness of the robot's representation of the environment. However, building highly detailed models of unstructured environments and performing symbol grounding in their context is computationally expensive and places a runtime bottleneck on the model described by Equation~\ref{eqn:basic-2}. A recent line of research~\cite{patki18a,patki2019a} has shown that perception can be adapted by leveraging language to build task-specific, compact world models for efficient symbol grounding. Figure~\ref{fig:system-architecture} shows the architecture of the proposed model that uses adaptive perception for the task of exploratory mobile manipulation in a priori unknown environments. We leverage adaptive perception online to build compact maps of the environment as the robot explores it. We hypothesize that using adaptive perception will improve the computational efficiency of semantic mapping (Eqn.~\ref{eqn:world-distribution}) and behavior inference (Eqn.~\ref{eqn:basic-1}), and thus improve the runtime language understanding (Eqn.~\ref{eqn:basic-2}). In the following sections, we describe each of the individual learned models of the proposed architecture.
\subsection{Adaptive Perception}

In practice, a large fraction of the objects and the corresponding symbols are inconsequential to inferring the meaning of an utterance. In such cases, there exists a compact environment representation that is sufficient to interpret the utterance. A recent line of work~\cite{patki18a,patki2019a} proposes adapting the robot's perception pipeline according to the demands of the language utterance. The goal is to quickly provide minimal task-relevant world models that are sufficiently expressive to permit accurate and fast language grounding. Following ~\cite{patki18a} we learn a Distributed Correspondence Graph (DCG)~\cite{paul2018efficient} based probabilistic model that exploits natural language in order to infer a small, succinct subset of perceptual classifiers $\bm{P}^*_t = f\left( \bm{P}, \bm{\Lambda}_t \right)$ as conditioned on the utterance $\bm{\Lambda}_t$. This allows dynamic adaptation of the robot's perceptual capabilities according to the current task resulting in compact models of the world $\Upsilon^*_t$.
\begin{figure}[!tb]
  \centering
  \subfigure[cluttered box]{
    \includegraphics[width=0.32\linewidth]{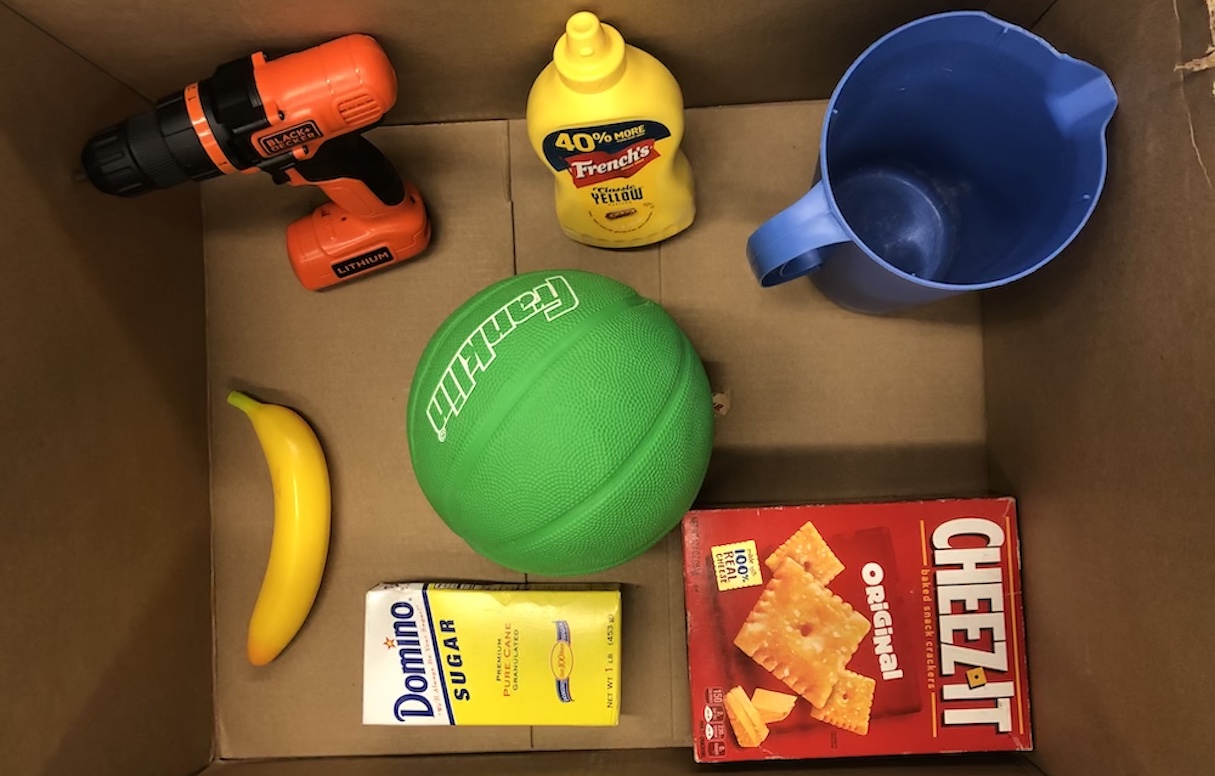}
  }\hfil
  \subfigure[exaustive perception]{
    \includegraphics[width=0.32\linewidth]{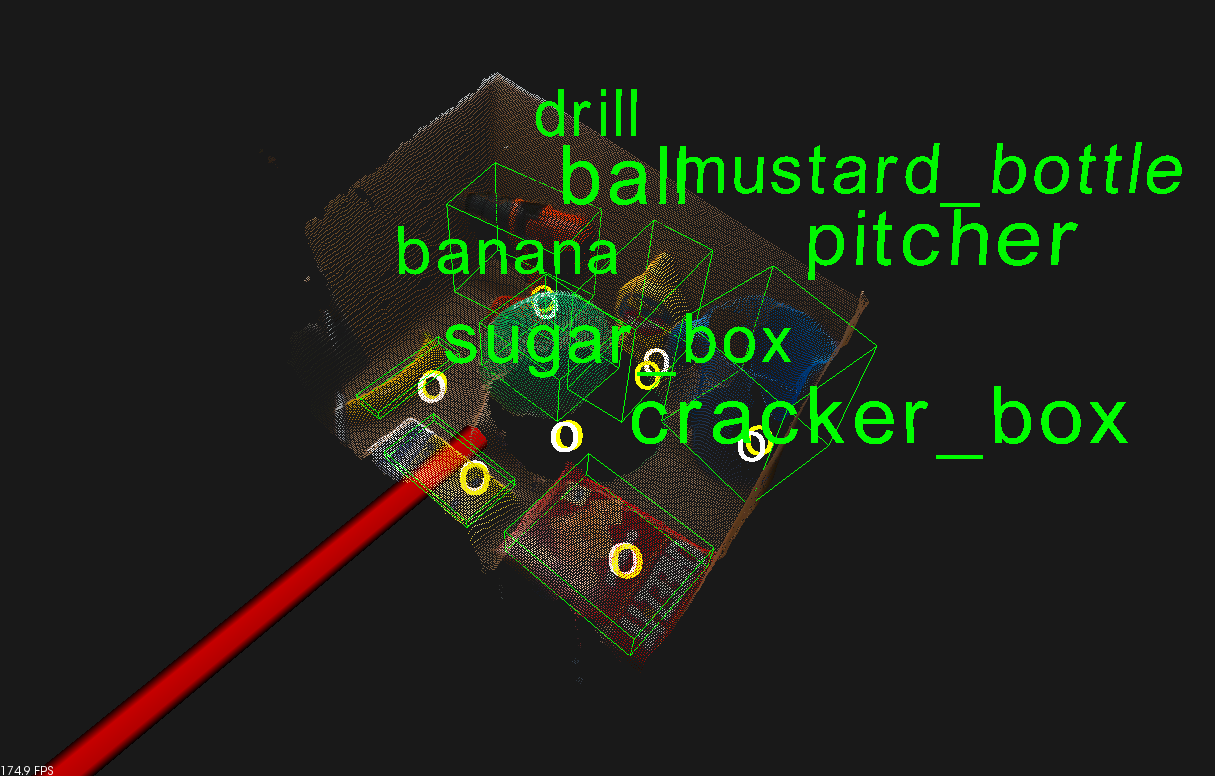}
  }\hfil
  \subfigure[adaptive perception]{
    \includegraphics[width=0.32\linewidth]{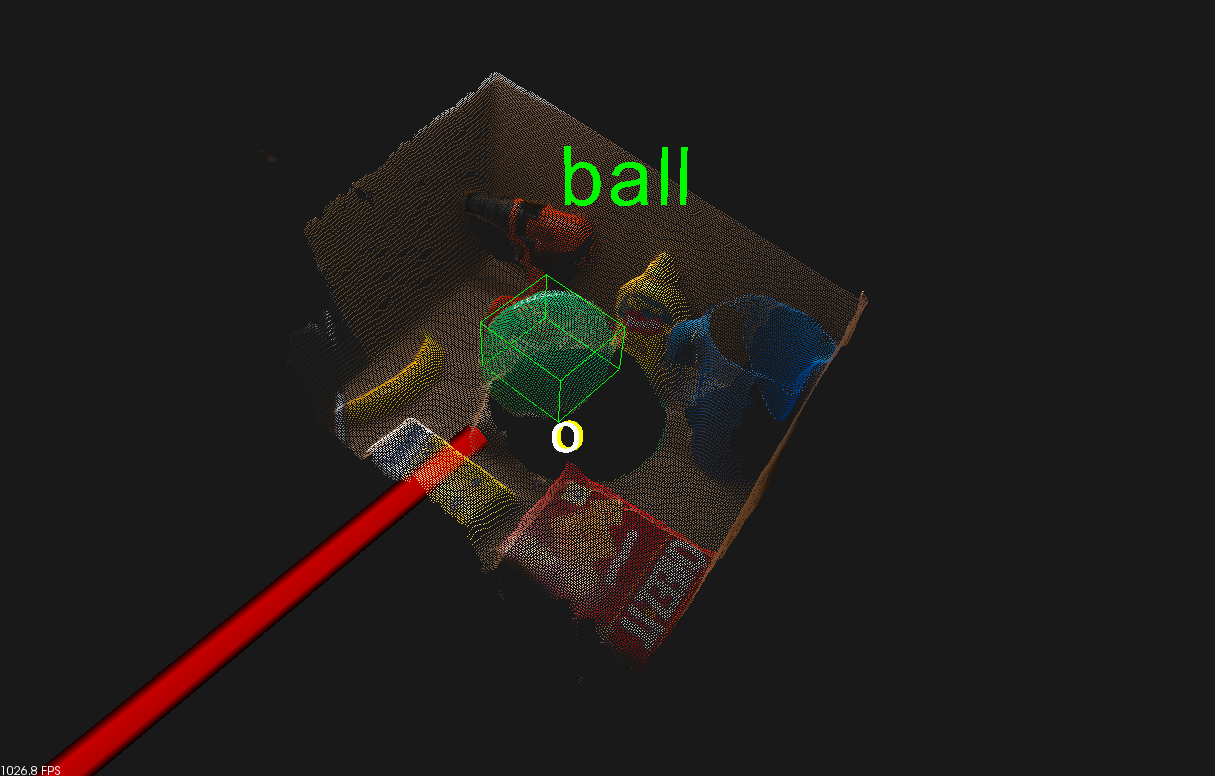}
  }
  \caption{A comparison of world models generated by (b) exhaustive and (c) adaptive perception after first observing a cluttered box for the instruction ``pick up the ball inside the box''.}
  \label{fig:ap_ep_comparison}
\end{figure}

Previously, adaptive perception has been shown to be effective in building compact models of the world from single observations~\cite{patki18a} or a log of past observations~\cite{patki2019a}. The proposed architecture leverages adaptive perception online in the context of SLAM to build compact maps of the novel environment during exploration.

\subsection{Semantic Mapping with Adaptive Perception}
We model the robot's environment as a semantic graph~\cite{walter13} $\bm{\Upsilon}_t = \{\bm{G}_t,\bm{X}_t,\bm{L}_t\}$. The topology $\bm{G}_t$ is comprised of nodes $n_i$ that represent distinct objects and locations and edges $e_{ij}$ that express spatial relationships between pairs of nodes (e.g., as inferred from language and the robot's motion). The metric map $\bm{X}_t$ associates a pose $x_i$ with each node $n_i$ in the graph in similar fashion to pose graph SLAM~\cite{kaess08}. The layer $\bm{L}_t$ expresses semantic attributes of each node (e.g., the type of each region/object and its colloquial name).

In order to follow instructions in a priori unknown environments, we leverage information about the environment implicit in a given utterance to maintain an informed distribution $\overline{\bm{\Upsilon}}_t$ over possible world models. We learn a DCG-based~\cite{paul2018efficient} probabilistic model that exploits natural language to infer a distribution over the available ``annotations'' $\bm{\alpha}_t$ (of which there may be none). These annotations include the type and relative location of different objects and regions in the environment. As an example, consider the instruction ``get the drill from the box''. DCG inference yields a distribution that assigns a high likelihood to annotations that suggest the existence of one or more objects of type ``box'' and ``drill''. High likelihood is associated with spatial relations that express a drill object as being ``inside'' a box object. The instruction can then be grounded in the context of a distribution of hypothesized worlds that incorporate the inferred annotations.

We maintain this distribution via a Rao-Blackwellized particle filter (RBPF)~\cite{doucet00, hahnel03a}, using a sample-based distribution over topologies, a Gaussian distribution over the metric map, and a Dirichlet over semantic information. We sample changes to the topology (as represented by a collection of particles) according to sensor-based observations and language-based annotations. We then update the resulting distribution over the metric map using an extended information filter. Maintaining a distribution over highly detailed world models and grounding instructions in their context is computationally non-trivial and places a bottleneck on the runtime efficiency of instruction following. We depart from previous work~\cite{hemachandra15} by integrating adaptive perception to maintain a distribution over compact maps $\overline{\bm{\Upsilon}}^*_t$ that afford more efficient behavior inference
\begin{equation}
  p(\overline{\bm{\Upsilon}}^*_t \vert \bm{\Lambda}_{1:t}, \bm{z}_{1:t}, \bm{u}_{1:t}, \bm{P}_t^*).
  \label{eqn:approximate-world-distribution}
\end{equation}

\subsection{Behavior Inference with Adaptive Perception}

We frame the problem of behavior inference as one of inferring a distribution over grounded behaviors $\overline{\bm{\beta}}_t$ for a given utterance in the context of the distribution over hypothesized worlds $\overline{\bm{\Upsilon}}_t$. When the space of symbols (groundings) $\bm{\Gamma}$ is large, the environment $\Upsilon$ is unstructured, and the free-form utterance $\bm{\Lambda}$ is complex and, making exact inference~\eqref{eqn:basic-1} becomes computationally intractable. Distributed Correspondence Graphs (DCG)~\cite{paul2018efficient} employ an approximate factorization of the grounding distribution in Equation~\ref{eqn:basic-1} that assumes conditional independence across the linguistic and symbolic constituents according to the hierarchical structure of language. DCG frames language understanding as an association problem by introducing the notion of correspondence variables $\phi_{ij} \in \bm{\Phi}$ that associate linguistic elements $\lambda_i \in \bm{\Lambda}$ (e.g., words and phrases) with symbols $\gamma_{ij} \in \bm{\Gamma}$.
In practice, a large fraction of the object and region symbols are irrelevant to inferring the meaning of an utterance. In such cases, there exists a compact environment representation $\Upsilon^{*}$ that is sufficient to interpret the utterance. Reasoning over compact world models reduces the size of the search space, improving the complexity of inference. Following our earlier work~\cite{patki18a,patki2019a}, we use adaptive perception to build these compact world representations. DCG inference then follows as a search for the correspondence variables $\bm{\Phi}^*$ that maximize the following factored distribution. Note that the grounding for a phrase depends on the child phrase groundings as contained in the true correspondence $\bm{\Phi}_{ci}$
\begin{equation}
  \bm{\Phi}^* = \argmax{\phi_{ij} \in \bm{\Phi}} \; \prod\limits_{ i = 1 }^{ \lvert \bm{\Lambda} \rvert } \prod\limits_{ j = 1 }^{ \lvert \bm{\Gamma} \rvert } p( \phi_{ij} \vert \gamma_{ij}, \lambda_i, \bm{\Phi}_{ci}, \Upsilon^*).
  \label{eqn:dcg-1}
\end{equation}
In a priori unknown environments, our framework performs grounding inference (Eqn.~\ref{eqn:dcg-1}) for each hypothesized world in the distribution. Thus the cost of behavior inference tends to be linear in the number of particles used to maintain the world distribution. Due to the computational advantage of performing inference over compact world models, our framework allows to reason over a larger distribution of possible worlds while being time efficient. Behavior inference finally yields a distribution over behaviors $\overline{\bm{\beta}}_t$. Each behavior $\beta^i_t \in \overline{\bm{\beta}}_t$ is parameterized by a type (navigate, retrieve or pickup) and a goal pose $g^i_t \in \bm{G}_t$.

\subsection{Planning Under Uncertainty}
We hypothesize that leveraging adaptive perception for semantic mapping and behavior inference will improve the runtime of the proposed system (Eqn.~\ref{eqn:proposed_model}) compared to our baseline (Eqn.~\ref{eqn:basic-2}).
\begin{equation}
  \bm{x}^*_t = \argmax{\bm{x}_t \in \bm{X}_t} \; \sum_{\Upsilon^{*i}_t \in \overline{\bm{\Upsilon}}^{*}_t} \underbrace{p(\bm{x}_t \vert \bm{\beta}^{i}_t, \Upsilon_t^{*i})}_{\text{path planning}} \times
  \underbrace{p( \bm{\beta}^{i}_t \vert \bm{\Lambda}_t, \Upsilon^{*i}_t)}_{\substack{\text{behavior inference} \\ \text{with adaptive perception}}} \times \underbrace{p( \Upsilon_t^{*i} \vert  \bm{\Lambda}_{1:t}, \bm{z}_{1:t}, \bm{u}_{1:t}, \bm{P}^*_t)}_{\substack{\text{semantic mapping} \\ \text{with adaptive perception}}}
  \label{eqn:proposed_model}
\end{equation}
Given a distribution over behaviors, identifying a suitable trajectory $\bm{x}^*_t$ amounts to a planning under uncertainty problem. We solve this problem with a policy that greedily chooses the best behavior $\beta_t^*$ that maximizes the following optimization function
\begin{equation}
  \beta_t^* = \argmax{\beta_t^i \in \overline{\bm{\beta}}_t} \;  \underbrace{\psi(\beta_t^i)}_{\substack{\text{decaying gaussian} \\ \text{cost function}}} \times \underbrace{p( \beta_t^{i} \vert \bm{\Lambda}_t, \Upsilon_t^{*i} )}_{\substack{\text{behavior inference} \\ \text{with adaptive perception}}} \times \underbrace{p( \Upsilon_t^{*i} \vert  \bm{\Lambda}_{1:t}, \bm{z}_{1:t}, \bm{u}_{1:t}, \bm{P}^*_t)}_{\substack{\text{semantic mapping} \\ \text{with adaptive perception}}},
  \label{eqn:proposed_model_2}
\end{equation}
where $\psi(\beta_t^i) = e^{ - d(r_t,g^i_t)^{2} / 10 }$ is a decaying Gaussian value function with $d(r_t,g^i_t)$ being the Euclidean distance between the robot and the goal location $g^i_t$ corresponding to the behavior $\beta_t^i$. This allows the robot to favor exploring goals that might be less likely but are closer to the robot. This process is repeated until the robot completes the instructed task.

%% file: experimental-design.tex

\section{Experimental Setup}\label{sec:experiments}

To evaluate the scalability of this framework, we performed experiments on a Clearpath Husky A200 Unmanned Ground Vehicle outfitted with a Universal Robotics UR5 arm and Robotiq 3-finger Adaptive Robot Gripper (Fig.~\ref{fig:motivation}). An Intel RealSense D435 RGB-D camera was mounted on the UR5 wrist was for object detection, while a Hokuyo UTM-30LX LIDAR was used for localization. The object detection pipeline consisted of YOLO V3-based~\cite{yolov3} object detector trained on the COCO dataset as well as 15 tiny YOLO V3 detectors each trained on individual classes from the Open Images V4~\citep{OpenImages} and YCB~\citep{calli2015benchmarking} datasets. The sensing range was limited to 4.5\,m indoors and 7\,m outdoors. The natural language understanding models were trained on a data-augmented corpus of approximately 115 instructions annotated separately for perception, behavior, and map inferences. The primary contribution of this work is an efficient approach to instruction following in unknown environments, and not the underlying grounding model itself, which has previously been shown to handle a large diversity of utterances~\citep{hemachandra11, paul2018efficient, hemachandra15, patki18a, patki2019a, tellex11, walter14b, walter13, hemachandra14}.

\begin{figure}[th!]
	\centering
	\subfigure[indoor environment]{\includegraphics[width=0.245\linewidth]{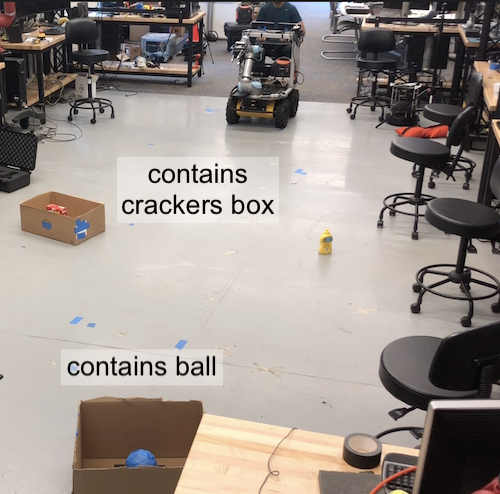}}\hfil%
	\subfigure[indoor world model]{\includegraphics[width=0.245\linewidth]{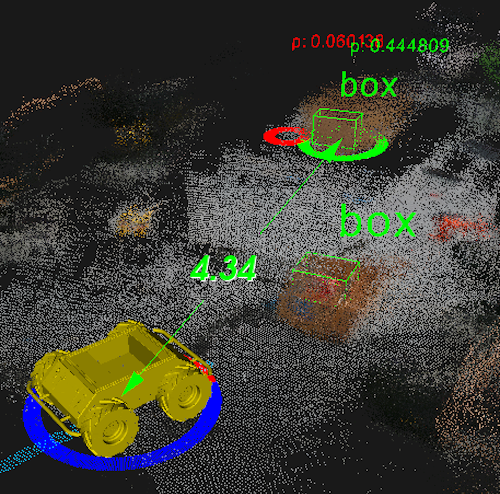}}\hfil%
	\subfigure[outdoor environment]{\includegraphics[width=0.245\linewidth]{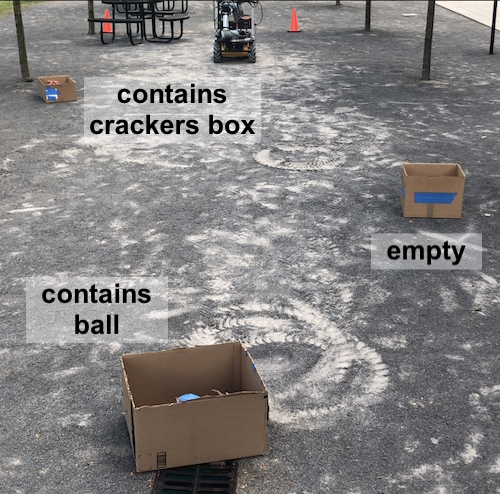}}\hfil%
	\subfigure[outdoor world model]{\includegraphics[width=0.245\linewidth]{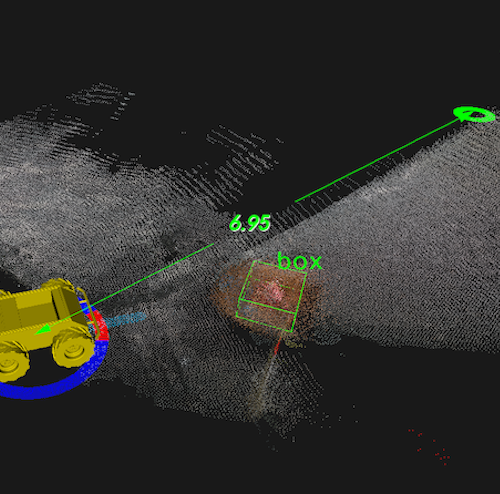}}
	\caption{Experiments were conducted in (a) indoor and (c) outdoor environments, and involved instructing the robot to retrieve an object from a box in a priori unknown environments. In both cases, the robot first explores the nearest box, which does not contain the target object. At this point, the robot either (b) explores the target box that comes within the robot's field-of-view (for the indoor experiments), or (d) further explores the environment (for the outdoor experiments). }
	\label{fig:experimental-enviroments}
\end{figure}

We designed the experiments to test the impact of adaptive perception (AP) on the runtime of instruction following in two different environmental settings. One of the experiments was designed as a controlled experiment in indoor settings to allow a direct time comparison of task execution times when using adaptive perception against exhaustive perception (EP). The other experiment was designed to demonstrate the exploratory capacity of the architecture. Figure \ref{fig:experimental-enviroments} illustrates the experimental setups. In both environments, the first command given was ``retrieve the ball inside the box''. The controlled experiment followed with a second command of ``pick up the crackers box inside the box'', whereas the exploratory experiment followed with ``go to the crackers box''. All commands in these experiments were provided as constituent parse trees. The box containing the ball was was not observable from the starting location, while the box with crackers box was observable.

To allow the robot to build accurate world models of its surroundings the robot's speed was limited to an average of 0.3\,m per perception cycle. Ten particles were used to maintain the world distribution for indoor experiments, while we used twenty particles for the outdoor experiments to account for the larger experimental workspace. A subset of past observations were stored during the execution of each behavior. When a second instruction was received during an adaptive perception experiment, a new set of perceptual classifiers was inferred and the semantic map was updated from those classifiers and stored observations.

While we motivate the problem in the context of large-scale unstructured environments, we focus on a detailed discussion of the above mentioned experiments to better convey the behavior of the framework. Additional experiments with more complex instructions and diverse environments is worthy of investigation and will be performed in the future. 

%% file: results.tex

\section{Results and Discussion} \label{sec:results}

\begin{figure}[!t]
  \centering
  \subfigure[Controlled EP Result] {
    \centering
    \includegraphics[width=0.485\linewidth]{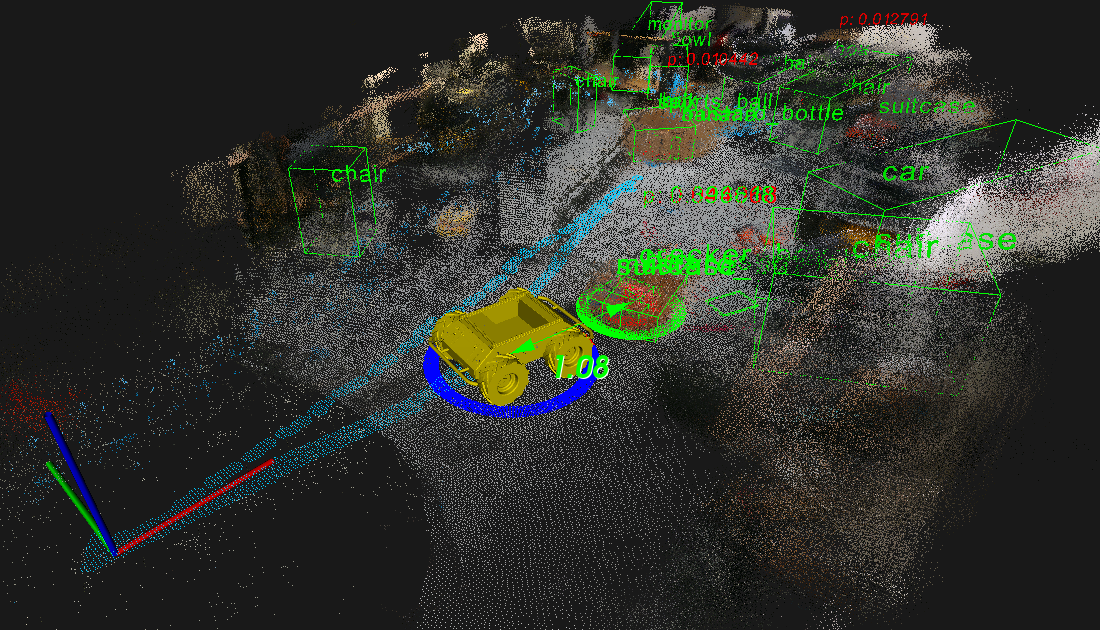}
  }\hfil%
  \subfigure[Controlled AP Result] {
    \centering
    \includegraphics[width=0.485\linewidth]{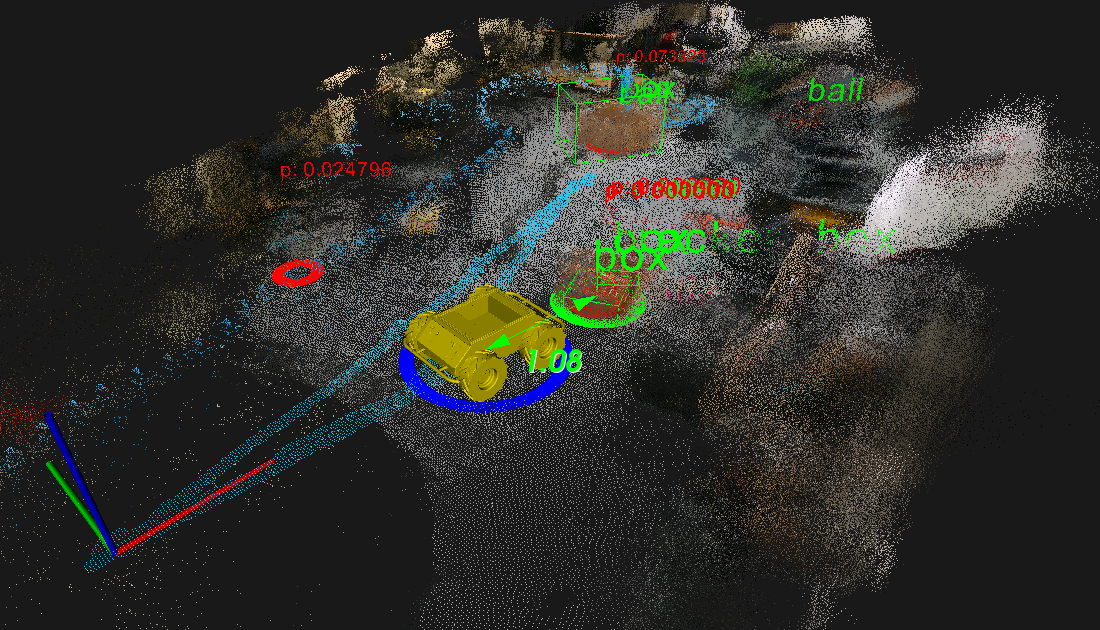}
  }\\
  \subfigure[Exploratory EP Result] {
    \centering
    \includegraphics[width=0.485\linewidth]{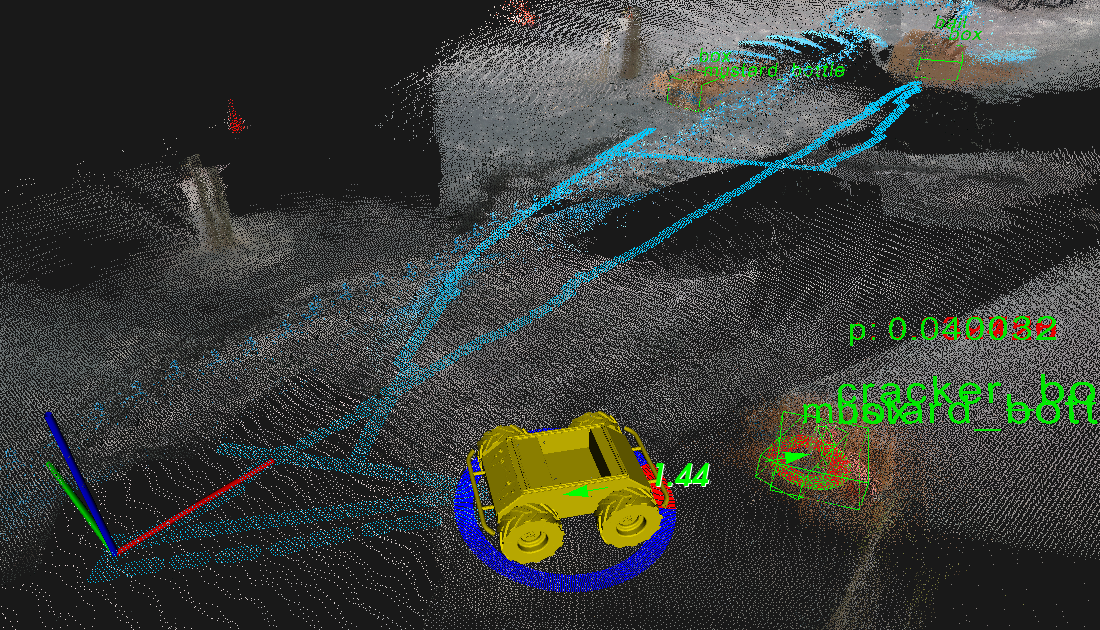}
  }\hfil%
  \subfigure[Exploratory AP Result] {
    \centering
    \includegraphics[width=0.485\linewidth]{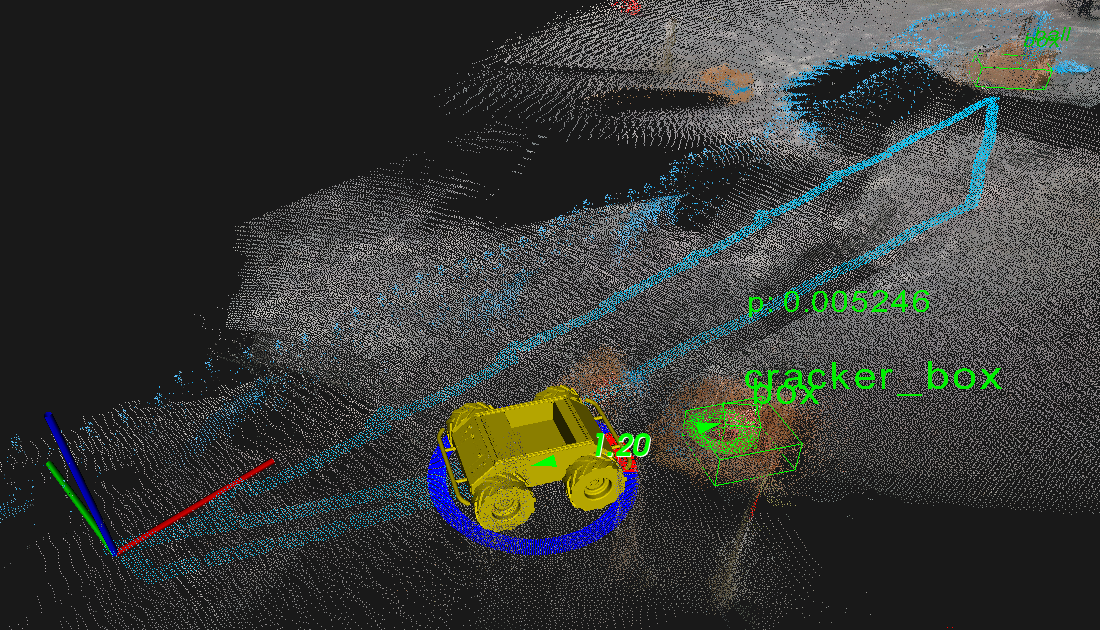}
  }
  \caption{The robot's visualization at the end of the experiment for each of the four trials run. The blue path illustrates the path that the robot took. The collection of green bounding boxes and their respective labels make up the world models generated during each trial. The first command issued in all trials was ``retrieve the ball inside the box'' and the second command issued was ``drive to the crackers box in the box''. }
  \label{fig:final-visualization}
\end{figure}

We evaluated the effect of adaptive perception on the runtime of various aspects of task execution as outlined in Table~\ref{table:results}. World models generated at the end of each trial are depicted in Figure~\ref{fig:final-visualization}. The impact of adaptive perception on the compactness of the generated world models is emphasized more for the indoor experiment as the indoor environment contained a higher number of objects (24 vs.\ 9). As behavior inference is performed on each hypothesized world model in the distribution, the efficiency gains provided by adaptive perception would enable reasoning over more number of environment hypotheses in the same amount of time. This is important as it allows for maintaining more particles and thus more efficient exploration.
\begin{table}[!h]
  \centering
  \begin{tabularx}{1.0\linewidth}{lXXXX}
      \toprule
      & AP \newline Controlled & EP \newline Controlled & AP \newline Exploratory & EP \newline Exploratory \\
      \midrule
    avg.\ behavior inf.\ time per world (s) & 0.020         & 0.035         & 0.016          & 0.019          \\
    avg.\ perception loop period (s)        & 0.700         & 4.141         & 0.655          & 4.099          \\
    time spent analyzing past obs.\ (s)    & 19.6          & 0             & 13.5           & 0              \\
    first task time (s)            & 186.6         & 351.2         & 214.4          & 593.5          \\
    second task time (s)           & 90.5          & 149.5         & 22.0           & 20.4           \\
    total detected objects         & 9             & 24            & 8              & 11             \\
  \bottomrule
  \end{tabularx}
  \vspace{2mm}
  \caption{Computational efficiency with adaptive (AP, ours) and exhaustive perception (EP).}
  \label{table:results}
  \vspace{-2mm}
\end{table}

The difference in behavior inference time is less prominent in the outdoor trials due to the the sparsity of the environment. The noticeable reduction in the perception run-time enables our framework to operate more efficiently while processing the same number of observations. This reduces the time required for task execution. However, an advantage of using exhaustive perception is that it does not require re-analyzing past observations when interpreting new instructions. Such is necessary with adaptive perception when subsequent instructions involve new detectors. This resulted in a shorter second task completion time for exhaustive perception in the outdoors environment.

%% file: conclusion.tex

\section{Conclusions} \label{sec:conclusions}

We proposed a novel framework that improves the efficiency of grounding natural language instructions in a priori unknown environments.  Integral to our approach is the coupling of three learned language understanding models with distinct symbolic representations for adaptive perception, map inference, and behavior inference.  Physical experiments on a mobile manipulator demonstrate higher language grounding efficiency over a contemporary baseline that employs exhaustive perception. In ongoing work, we are exploring hierarchical spatial-semantic representations and more complex mobile manipulation tasks that consider affordances and dynamics of perceived objects.